\documentclass[letterpaper, 10 pt, conference]{ieeeconf}  

\IEEEoverridecommandlockouts                              

\usepackage{graphics} 
\usepackage{epsfig} 
\usepackage{times} 
\usepackage{amsmath} 
\usepackage{amssymb}  
\usepackage{cite}
\usepackage{bm}
\usepackage{float}

\usepackage[symbol]{footmisc}

\title{\LARGE \bf
On the Modularity of Elementary Dynamic Actions
}
 
\author{Moses C. Nah,$^{1}$ Johannes Lachner,$^{1,2}$ Federico Tessari,$^{1}$ and Neville Hogan$^{1,2}$
\thanks{$^{1}$Department of Mechanical Engineering, Massachusetts Institute of Technology, Cambridge, MA, USA.}
\thanks{$^{2}$Department of Brain and Cognitive Sciences, Massachusetts Institute of Technology, Cambridge, MA, USA.}
\thanks{ This work was supported in part by the MIT/SUSTech Centers for Mechanical Engineering Research and Education. MCN was supported in part by a Mathworks Fellowship. JL was supported in part by the MIT-Novo Nordisk Artificial Intelligence Postdoctoral Fellows Program. 
}
\thanks{This work has been submitted to the IEEE for possible publication. Copyright may be transferred without notice, after which this version may no longer be accessible.
}
}

\begin{document}

\maketitle
\thispagestyle{empty}
\pagestyle{empty}

\begin{abstract}
In this paper, a kinematically modular approach to robot control is presented. 
The method involves structures called Elementary Dynamic Actions and a network model combining these elements. 
With this control framework, a rich repertoire of movements can be generated by combination of basic modules. 
The problems of solving inverse kinematics, managing kinematic singularity and kinematic redundancy are avoided. 
The modular approach is robust against contact and physical interaction, which makes it particularly effective for contact-rich manipulation.
Each kinematic module can be learned by Imitation Learning, thereby resulting in a modular learning strategy for robot control. 
The theoretical foundations and their real robot implementation are presented. 
Using a KUKA LBR iiwa14 robot, three tasks were considered: (1) generating a sequence of discrete movements, (2) generating a combination of discrete and rhythmic movements, and (3) a drawing and erasing task.
The results obtained indicate that this modular approach has the potential to simplify the generation of a diverse range of robot actions.
\end{abstract}

\section{INTRODUCTION}
To generate complex motor behavior that can match that of humans, robot control based on motor primitives has been proposed \cite{schaal1999imitation, ijspeert2002movement, calinon2007active, ijspeert2013dynamical, fang2019survey, hogan2012dynamic,hogan2013dynamic ,hogan2017physical,hogan2018impedance,saveriano2023dynamic}.
The method originates from motor neuroscience research, where the complex motor behavior of biological systems appears to be generated by a combination of fundamental building blocks \cite{sherrington1906integrative,bernstein1935problem,morasso1981spatial,flash2005motor,bizzi2017acquisition,d2016modularity}.
By parameterizing a controller using motor primitives, robots can efficiently learn, adapt, and execute a wide range of tasks. 

In robotics, two distinct motor-primitive approaches have been identified: Elementary Dynamic Actions (EDA)\footnote[1]{The original name suggested by Hogan and Sternad \cite{hogan2012dynamic} was ``Dynamic \emph{Motor} Primitives.'' However, to avoid confusion due to the similarity to ``Dynamic \emph{Movement} Primitives,'' we instead use ``Elementary Dynamic Actions.''} \cite{hogan2012dynamic,hogan2013dynamic,hogan2017physical} and Dynamic Movement Primitives (DMP) \cite{ijspeert2013dynamical,schaal2006dynamic,saveriano2023dynamic}.
EDA provides a modular framework for robot control that also accounts for physical interaction \cite{hogan2012dynamic,hogan2013dynamic, hogan2017physical,hogan2018impedance,hogan2022contact}. 
One of its applications, impedance control \cite{hogan1985impedance}, has been a prominent approach for tasks involving contact and physical interaction. 
DMP provides a rigorous mathematical framework to generate movements of arbitrary complexity \cite{ijspeert2013dynamical}. 
Its prominent application, Imitation Learning (or Learning from Demonstration \cite{calinon2007active}), provides a systematic method to learn (or imitate) trajectories that are provided by demonstration. 

Although both EDA and DMP have provided useful frameworks for robot control, the potential advantages of integrating these two approaches have not yet been thoroughly explored.
EDA enhances modularity in motion planning and robot command execution, which greatly simplifies robot control. Nevertheless, programming at the kinematic level remains challenging \cite{hogan1985impedance, lachner2022geometric}. Therefore, incorporating learning-based methods, such as Imitation Learning, could significantly improve the usability of this approach.
DMP is a prominent method to generate a rich repertoire of movements. Yet, mapping these movements to robot commands is not trivial and requires additional consideration. For DMP trajectories generated in task-space, additional methods must be included to map these trajectories to joint position or torque commands, e.g., managing kinematic singularities \cite{buss2005selectively} and kinematic redundancy \cite{nakamura1986inverse} of the robot. Therefore, merging DMP with the modularity of EDA will facilitate robot control.

In this paper, we combine EDA and DMP to achieve a modular learning approach for robot control. We show how a wide range of robot movements can be produced by combining basic modules. This has the potential to facilitate the programming and control of more difficult robot tasks. The approach presented preserves the advantages of EDA for tasks involving contact and physical interaction. Hence, the approach can be employed for contact-rich manipulation. 

We demonstrate our approach with an implementation on a real robot, using a KUKA LBR iiwa for modular task-space control. These demonstrations illustrate how the proposed approach can simplify the generation of a range of different robot tasks.
We present three different combinations of modules: (1) a sequence of discrete movements, (2) a combination of discrete and rhythmic movements, and (3) a combination of rhythmic movements with Imitation Learning for a drawing and erasing task. Examples (1) and (2) highlight the kinematic modularity offered by EDA. 
Example (3) shows how the kinematic programming with EDA can be enhanced by imitation learning. 
The task also includes contact and physical interaction to showcase EDA's robustness in dealing with physical interaction. 

\section{Theoretical Foundations}
For the remainder of the paper, a torque-actuated $n$ degrees of freedom (DOFs) open-chain robotic manipulator will be considered. How to use the approach with position-actuated robots will briefly be discussed in Section \ref{sec:discussion_and_conclusion}.

\subsection{Elementary Dynamic Actions and the Norton Equivalent Network Model}\label{subsec:EDA_and_Norton_Network}
\begin{figure}[H]
    \centering
    \includegraphics[trim={3.0cm 8.5cm 4.0cm 2.0cm}, width=0.94\columnwidth, clip, page=1]{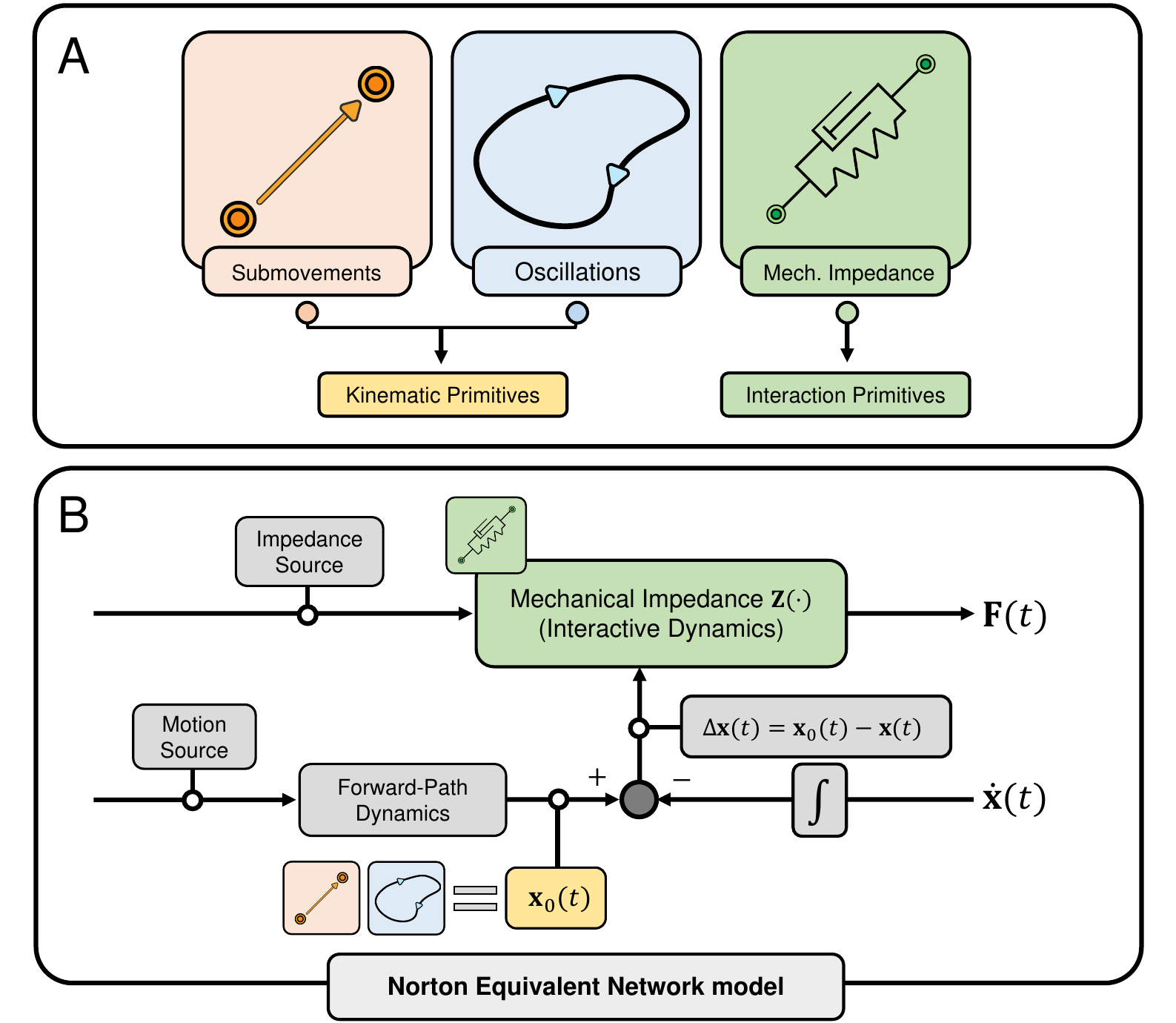}
    \caption{(A) Three Elementary Dynamic Actions (EDA). Submovements (orange box) and oscillations (blue box) correspond to kinematic primitives and mechanical impedances (green box) manage physical interaction. (B) EDA combined using a Norton equivalent network model. The virtual trajectory $\mathbf{x}_0(t)$ (yellow box) consists of submovements and/or oscillations, and mechanical impedance $\mathbf{Z}$ (green box) regulates interactive dynamics.}
    \label{fig:edas_w_Norton_Network}       
\end{figure}

EDA, introduced by Hogan and Sternad \cite{hogan2012dynamic,hogan2013dynamic,hogan2017physical}, consist of (at least) three distinct classes of primitives (Figure \ref{fig:edas_w_Norton_Network}A):
\begin{itemize}
    \item Submovements for discrete movements \cite{hogan2007rhythmic}.
    \item Oscillations for rhythmic movements \cite{hogan2007rhythmic}.
    \item Mechanical impedances to manage physical interaction \cite{hogan2018impedance}. 
\end{itemize}
Submovements and oscillations constitute the kinematic primitives, and mechanical impedance constitutes the interactive primitives of EDA.

\subsubsection{Kinematic Primitives --- Submovements and Oscillations}\label{subsubsec:EDA_Kinematic_Primitives}
A submovement $\mathbf{x}_0:\mathbb{R}_{\ge 0}\rightarrow \mathbb{R}^{n}$ is a smooth trajectory with its time derivative defined by:
\begin{equation}\label{eq:EDA_submovement}
    \dot{\mathbf{x}}_0(t) = \mathbf{v} \ \hat{\sigma}(t)
\end{equation}
In this equation, $t\in\mathbb{R}_{\ge 0}$ is time; $\hat{\sigma}:\mathbb{R}_{\ge 0}\rightarrow \mathbb{R}$ denotes a smooth unimodal basis function in which the integration over the time domain is 1, i.e., $\int_{0}^{\infty}\hat{\sigma}(t)dt=1$; 
$\mathbf{v}\in\mathbb{R}^{n}$ is the velocity amplitude array. 

Since submovements represent discrete movement, $\hat{\sigma}(t)$ has a finite support, i.e., $\exists T>0: \hat{\sigma}(t)=0$ for $t\ge T$.

Given an initial position $\mathbf{x}_0(t=0)\equiv \mathbf{x}_{i}\in\mathbb{R}^{n}$ and goal location $\mathbf{x}_0(t=T)\equiv \mathbf{x}_{g}\in \mathbb{R}^{n}$, $\mathbf{v}=\mathbf{x}_{g}-\mathbf{x}_{i}$.
Hence, submovement $\mathbf{x}_0(t)$ is defined by:
\begin{equation}\label{eq:EDA_submovement_integrated}
    \mathbf{x}_0(t) = \mathbf{x}_{i} + (\mathbf{x}_{g} - \mathbf{x}_{i}) f_{\sigma}(t)
\end{equation}
In this equation, $f_{\sigma}:\mathbb{R}_{\ge 0}\rightarrow \mathbb{R}_{\ge 0}$ is an integral of $\hat{\sigma}(t)$, i.e., $f_\sigma(t) = \int_{0}^{t} \hat{\sigma}(s)ds$.
Accounting for the definition of $\hat{\sigma}(t)$, $\forall t\ge T: f_{\sigma}(t)=1$.

An oscillation $\mathbf{x}_0:\mathbb{R}_{\ge 0}\rightarrow \mathbb{R}^{n}$ is a smooth non-zero trajectory which is a periodic function:
\begin{equation}\label{eq:EDA_oscillation}
    \forall t >0: ~~ \exists T>0: ~~ \mathbf{x}_0(t) = \mathbf{x}_0(t+T)
\end{equation}
Compared to submovements, oscillations represent rhythmic and repetitive motions.

\subsubsection{Interactive Primitive --- Mechanical Impedances}
Mechanical impedance $\mathbf{Z}:\mathbb{R}^{n}\rightarrow \mathbb{R}^{n}$ is an operator which maps (generalized) displacement $\Delta \mathbf{x}(t) \in \mathbb{R}^{n}$ to (generalized) force $\mathbf{F}(t)\in\mathbb{R}^{n}$ \cite{hogan2017physical, hogan2018impedance, hogan2022contact}:
\begin{equation*}
    \mathbf{Z}: \Delta \mathbf{x}(t) \longrightarrow \mathbf{F}(t)
\end{equation*}
In this equation, $\Delta \mathbf{x}(t)$ is the displacement of an actual trajectory of (generalized) position $\mathbf{x}(t)$ from a virtual trajectory $\mathbf{x}_0(t)$ to which the mechanical impedance is connected, i.e., $\Delta \mathbf{x}(t)=\mathbf{x}_0(t)-\mathbf{x}(t)$.
Loosely speaking, mechanical impedance is a generalization of stiffness to encompass the dynamic relation of force to displacement and its derivatives.

The definition of generalized displacement $\Delta \mathbf{x}(t)=\mathbf{x}_0(t)-\mathbf{x}(t)$ accounts for the actual space in which $\mathbf{x}(t)$ resides. 
For instance, if one considers the displacements, $\Delta \mathbf{p}(t)\equiv\mathbf{p}_0(t)-\mathbf{p}(t)$, where $\mathbf{p}_0(t), \mathbf{p}(t)\in \mathbb{R}^{3}$ are the virtual and actual end-effector position.
If one considers the displacement in $\text{SO}(3)$, $\Delta \mathbf{R}(t)\equiv\mathbf{R}(t)^{\top}\mathbf{R}_0(t)$, where $\mathbf{R}(t),\mathbf{R}_{0}(t)\in\text{SO}(3)$ are the virtual and actual end-effector orientation of the robot (Section \ref{subsec:robot_controller}).

Compared to the kinematic primitives, mechanical impedance is an interactive primitive which regulates the dynamics of physical interaction.
For instance, tactile exploration and manipulation of fragile objects should evoke the use of low stiffness, while tasks such as drilling a hole on a surface require high stiffness for object stabilization \cite{hogan2018impedance}.
Parameterizing control based on mechanical impedances provides beneficial stability properties, making this approach effective for tasks involving contact and physical interaction \cite{hogan1985impedance,lachner2022geometric}. 

Another feature of mechanical impedances is that they can be linearly superimposed even though each mechanical impedance is a nonlinear operator. 
This is the ``superposition principle of mechanical impedances'' \cite{hogan1985impedance,hogan2017physical,hogan2018impedance,hogan2022contact}:
\begin{equation}\label{eq:superposition_of_mechanical_impedances}
    \mathbf{Z} = \sum \mathbf{Z}_i  
\end{equation}
Note that the impedance operators include transformation maps (i.e., Jacobian matrices) (Section \ref{sec:method}).
This property simplifies a wide range of control tasks, as it provides a modular framework for robot control \cite{andrews1983impedance, newman1987high,hermus2021exploiting,hogan2017physical,hogan2022contact,hjorth2020energy}.

\subsubsection{Norton Equivalent Network Model}\label{subsubsec:Norton_network_model}
The three primitives of EDA can be combined using a Norton equivalent network model \cite{hogan2017physical}, which provides an effective framework to relate the three elements of EDA (Figure \ref{fig:edas_w_Norton_Network}B). 
The forward-path dynamics specifies the virtual trajectory $\mathbf{x}_0(t)$, which consists of submovements and/or oscillations. 
The mechanical impedance $\mathbf{Z}$, determines $\mathbf{F}(t)$ from $\Delta\mathbf{x}(t)$ which is eventually mapped to the robot joint torque command.
Hence, a key objective of EDA is to find appropriate choices of $\mathbf{x}_0(t)$ and $\mathbf{Z}$ to generate the desired robot behavior. 

The Norton equivalent network model clearly distinguishes the actual trajectory $\mathbf{x}(t)$ from the virtual trajectory $\mathbf{x}_0(t)$ to which the mechanical impedance is connected.
While $\mathbf{x}(t)$ is determined by the interaction with the environment (i.e., \textit{bidirectional}), the virtual trajectory $\mathbf{x}_0(t)$ can be chosen independently with respect to the environment  (i.e., \textit{unidirectional}) \cite{hogan2017physical}.
This allows submovements and/or oscillations to be directly superimposed at the level of virtual trajectory $\mathbf{x}_0(t)$.
Not only submovements and/or oscillations, but any trajectory generating methods such as DMP can be seamlessly included to generate $\mathbf{x}_0(t)$:
\begin{equation}\label{eq:virtual_trajectory_summations}
    \mathbf{x}_0(t) = \sum \mathbf{x}_{0,sub}(t) + \sum \mathbf{x}_{0,osc}(t) + \sum \mathbf{x}_{0,dmp}(t)
\end{equation}
In this equation, $\mathbf{x}_{0,sub}(t)$, $\mathbf{x}_{0,osc}(t)$, $\mathbf{x}_{0,dmp}(t)$ are submovement, oscillation, and trajectory generated by DMP, respectively.
As can be seen in the equation, this concept provides kinematic modularity which is capable of simplifying the generation of a wide range of movements. 

\subsection{Dynamic Movement Primitives and Imitation Learning}\label{subsec:DMP_and_IL}
DMP, introduced by Ijspeert, Schaal, et al.\cite{ijspeert2013dynamical} consists of three distinct classes of primitives: canonical system, nonlinear forcing term, and transformation system. 
Using these three primitives, DMP can generate both discrete and rhythmic movements.
For this overview, we focus on DMP for discrete movement, although the generalization to rhythmic movement is straightforward for the application \cite{ijspeert2013dynamical}.
In this paper, DMP is used to generate the virtual trajectory of EDA. 

\subsubsection{Dynamic Movement Primitives}
A canonical system for discrete movement $s:\mathbb{R}_{\ge 0}\rightarrow \mathbb{R}_{\ge 0}$ is a scalar variable governed by a stable first-order differential equation \cite{saveriano2023dynamic}:
\begin{equation*}
    \tau \dot{s}(t) = -\alpha_s s(t)
\end{equation*}
In this equation, $\alpha_s, \tau \in \mathbb{R}_{> 0}$, where $\tau$ is the duration of the discrete movement.

A nonlinear forcing term for discrete movement, $\mathbf{F}_{s}:\mathbb{R}_{\ge 0}\rightarrow \mathbb{R}^{n}$, which takes the canonical system $s(t)$ as the function argument, is defined by:
\begin{align*}
\begin{split}
    \mathbf{F}_{s}(s(t)) &= \frac{\sum_{i=1}^{N}\mathbf{w}_i\phi_i(s(t))}{\sum_{i=1}^{N}\phi_i(s(t))} s(t) \\
    \phi_i(s(t)) &= \exp\big\{ -h_i(s(t)-c_i)^2 \big\}
\end{split}
\end{align*}
In these equations, $\phi_i:\mathbb{R}_{\ge 0}\rightarrow \mathbb{R}_{\ge 0}$ is the $i$-th basis function of the nonlinear forcing term which is a Gaussian function;
$N$ is the number of basis functions; $\mathbf{w}_i\in\mathbb{R}^{n}$ is the weight array and $c_i$, $h_i$ determine the center and width of the $i$-th basis function, respectively.

The nonlinear forcing term can be concisely denoted by:
\begin{equation*}
    \mathbf{F}_{s}(s(t)) = \mathbf{W}\mathbf{a}(t), ~~~~ a_i(t) = \frac{\phi_i(s(t))}{\sum_{i=1}^{N} \phi_i(s(t))}s(t)
\end{equation*}
In this equation, $\mathbf{W}\in\mathbb{R}^{n\times N}$ is the weight matrix with $\mathbf{w}_i$ as the $i$-th column; $a_i(t)$ is the $i$-th element of $\mathbf{a}(t)\in\mathbb{R}^{N}$.
 
A transformation system is a collection of $n$ second-order differential equations with a scaled nonlinear forcing term as an input:
\begin{equation*}
\resizebox{.9\hsize}{!}{$
    \begin{aligned}
        \tau \dot{\mathbf{x}}_0(t) &= \mathbf{z}(t) \\
        \tau \dot{\mathbf{z}}(t) &= \alpha_z \{ \beta_z (\mathbf{x}_{g}-\mathbf{x}_0(t)) -\mathbf{z}(t)\} + \text{diag}(\mathbf{x}_{g}-\mathbf{x}_{i}) \mathbf{F}_{s}(s(t))
    \end{aligned}  
    $}
\end{equation*}
In these equations, $\alpha_z, \beta_z\in \mathbb{R}_{>0}$; $\mathbf{z}(t)\in\mathbb{R}^{n}$ denotes the time-scaled velocity of $\mathbf{x}_0(t)$; $\text{diag}: \mathbb{R}^{n}\rightarrow \mathbb{R}^{n\times n}$ constructs a diagonal matrix with its elements defined by its array argument. 

Given the initial conditions $\mathbf{x}_0(t=0)\equiv \mathbf{x}_i$, $\mathbf{z}(t=0)=\mathbf{0}$ and the nonlinear forcing term $\mathbf{F}_s$, the differential equation for the transformation system is forward integrated to generate $\mathbf{x}_0(t)$.
Without $\mathbf{F}_s$, trajectory $\mathbf{x}_0(t)$ follows a response of a stable second-order linear system which converges to $\mathbf{x}_g$.
To generate a wider range of movements, the weights of the nonlinear forcing term $\mathbf{F}_s$ are learned through various methods \cite{kober2013reinforcement,saveriano2023dynamic}. 
One of the prominent methods is Imitation Learning, which learns the weight array $\mathbf{W}$ by demonstration.

\subsubsection{Imitation Learning}\label{subsubsec:imitation_learning}
Imitation Learning generates (or mimics) a demonstrated trajectory $\mathbf{x}^{(d)}(t)$ from the transformation system by learning the best-fit weights $\mathbf{W}^{*}$.
Given $P$ samples of $(\mathbf{x}^{(d)}(t_j),$ $\dot{\mathbf{x}}^{(d)}(t_j)$, $\ddot{\mathbf{x}}^{(d)}(t_j))$ for $j\in[1,2,\cdots, P]$, the best-fit weights are calculated using Linear Least Square Regression \cite{saveriano2023dynamic}:
\begin{align*}
\begin{split}
    \mathbf{W}^{*} &= \mathbf{B}\mathbf{A}^{\top} (\mathbf{A}\mathbf{A}^{\top})^{-1}  \\
    \mathbf{A} &= 
    \begin{bmatrix}
        \mathbf{a}(t_1) & \mathbf{a}(t_2) & \cdots & \mathbf{a}(t_{P})     
    \end{bmatrix} \\
    \mathbf{B} &= \{\text{diag}(\mathbf{x}_{g}^{(d)}-\mathbf{x}_i^{(d)})\}^{-1}
    \begin{bmatrix}
        \mathbf{b}(t_1) & \mathbf{b}(t_2) & \cdots & \mathbf{b}(t_{P}) 
    \end{bmatrix} \\
    \mathbf{b}(t_j) &= \tau^{2}\ddot{\mathbf{x}}^{(d)}(t_j) + \alpha_z \tau \dot{\mathbf{x}}^{(d)}(t_j) + \alpha_z \beta_z ( \mathbf{x}^{(d)}(t_j) - \mathbf{x}_g^{(d)} )
\end{split}    
\end{align*}     
In these equations, $\mathbf{A}\in\mathbb{R}^{N\times P}$, $\mathbf{B}\in\mathbb{R}^{n\times P}$.
For the initial and goal positions of the demonstrated trajectory, $\mathbf{x}_{i}^{(d)}$ and $\mathbf{x}_g^{(d)}$, the first and last samples are used, i.e., $\mathbf{x}^{(d)}(t_1)=\mathbf{x}_i^{(d)}$ and $\mathbf{x}^{(d)}(t_P)=\mathbf{x}_g^{(d)}$.

Along with Linear Least Square Regression, Locally Weighted Regression can also be used to find the best-fit weights \cite{atkeson1997locally,ijspeert2013dynamical}. 
However, for a higher accuracy of imitating $\mathbf{x}^{(d)}(t)$, Linear Least Square Regression is used. 

\section{The Three Control Tasks and Methods}\label{sec:method}
In this section, we introduce three control tasks and the corresponding methods to achieve them.
\begin{enumerate}
    \item Generating a sequence of discrete movements.
    \item Generating a combination of discrete and rhythmic movements.
    \item Drawing and erasing a path on a table. 
\end{enumerate}
The first two tasks highlight the kinematic modularity of EDA. 
The final task provides an illustrative example of combining Imitation Learning of DMP with EDA.
The task also involves physical contact which illustrates the benefit of EDA for contact-rich manipulation.
All three tasks consider task-space control, both position and orientation, using a kinematically redundant robot, i.e., $n> 6$.

\subsection{The Robot Controller}\label{subsec:robot_controller}
The torque command of a robot, $\bm{\tau}_{in}(t)\in\mathbb{R}^{n}$ is defined by superimposing three mechanical impedances (Eq.~\eqref{eq:superposition_of_mechanical_impedances}):
\begin{equation*}
    \bm{\tau}_{in}(t) = \mathbf{Z}_q(\mathbf{q}(t)) + \mathbf{J}_{p}^{\top}(\mathbf{q})\mathbf{Z}_p( \Delta \mathbf{p}(t)) + \mathbf{J}_{r}^{\top}(\mathbf{q})\mathbf{Z}_{r}( \Delta \mathbf{R}(t))
\end{equation*}
where:
\begin{align}\label{eq:impedances}
    \begin{split}
    \mathbf{Z}_q(\mathbf{q}(t)) &= -\mathbf{B}_q \dot{\mathbf{q}}(t)\\ 
    \mathbf{Z}_p(\Delta \mathbf{p}(t)) &= \mathbf{K}_p \Delta \mathbf{p}(t) + \mathbf{B}_p  \Delta \dot{\mathbf{p}}(t) \\  
    \mathbf{Z}_r(\Delta \mathbf{R}(t)) &= \mathbf{K}_{r} \mathbf{R}(t) \mathbf{Log} (\Delta \mathbf{R}(t)) - \mathbf{B}_{r}
        \bm{\omega}(t)
    \end{split}
\end{align}
In these equations, $\mathbf{Z}_{q}$, $\mathbf{Z}_{p}$, $\mathbf{Z}_{r}$ denote mechancal impedances for joint-space, task-space position and task-space orientation, respectively; $\mathbf{q}\equiv \mathbf{q}(t)\in\mathbb{R}^{n}$ denotes the robot's joint trajectories;
$\mathbf{J}_{p}(\mathbf{q}), \mathbf{J}_{r}(\mathbf{q})\in\mathbb{R}^{3\times n}$ denote the Jacobian matrices for the translational velocity $\dot{\mathbf{p}}(t)$ and rotational velocity $\bm{\omega}(t)$ of the robot's end-effector, respectively, i.e., $\dot{\mathbf{p}}(t)=\mathbf{J}_{p}(\mathbf{q})\dot{\mathbf{q}}(t)$ and $\bm{\omega}(t)=\mathbf{J}_{r}(\mathbf{q})\dot{\mathbf{q}}(t)$;
$\mathbf{Log}:\text{SO}(3)\rightarrow \mathbb{R}^3$ denotes the Matrix Logarithm Map \cite{murray1994mathematical,lynch2017modern};
end-effector position $\mathbf{p}(t)$ and orientation $\mathbf{R}(t)$ are derived by the Forward Kinematics Map of the robot.

To generate goal-directed discrete movements that (1) maintain stability against contact and physical interaction and (2) manage kinematic redundancy, it is convenient to select positive definite stiffness matrices $\mathbf{K}_{p}, \mathbf{K}_{r}\in\mathbb{R}^{3\times 3}$ and damping matrices $\mathbf{B}_{p}, \mathbf{B}_{r}\in\mathbb{R}^{3\times 3}$, $\mathbf{B}_q\in\mathbb{R}^{n\times n}$ \cite{lachner2022geometric}.

It is worth emphasizing that the controller based on EDA only involves the Jacobian transpose map, but not its (generalized)-inverse. 
Hence, the controller avoids the problem of solving inverse kinematics, thus managing kinematic singularity and redundancy. 
Moreover, the controller does not involve local representations of $\text{SO}(3)$ (e.g., Euler angles), since it directly uses the spatial rotation matrices $\mathbf{R}\in \text{SO}(3)$. 
Hence, the controller avoids the problem of representation singularities \cite{jeong2022memory}. 
These features significantly simplify control in task-space. 

\subsection{Generating a Sequence of Discrete Movements }\label{subsec:sequence_of_discrete_movements}
Given an initial end-effector position $\mathbf{p}_{1}\in\mathbb{R}^3$, let $\mathbf{p}_{2}\in\mathbb{R}^3$ be a goal location which the robot's end-effector aims to reach.
This goal-directed discrete movement $\mathbf{p}(t)\rightarrow \mathbf{p}_{2}$ can be achieved by setting $\mathbf{p}_0(t)$ as $\mathbf{p}_{0,sub1}(t)$ \cite{takegaki1981new,slotine1991applied,arimoto2005natural}:
\begin{equation*}
    \mathbf{p}_{0,sub1}(t) = 
    \begin{cases}
      \mathbf{p}_1  &  0 \le t < t_i \\
      \mathbf{p}_{1} + ( \mathbf{p}_{2} - \mathbf{p}_{1}) f_{\sigma,1}(t) & t_i \le t < t_i + T_1 \\
      \mathbf{p}_{2} &  t_i + T_1 \le t
    \end{cases} 
\end{equation*}
In these equations, $f_{\sigma,1}(t)$ is a submovement starting at time $t_i$ with duration $T_1$ (Eq.~\eqref{eq:EDA_submovement}, \eqref{eq:EDA_submovement_integrated})

At time $t_g$, the goal location suddenly changes to a new goal location $\mathbf{p}_{3}\in\mathbb{R}^3$, which necessitates a second movement.
Using the kinematic modularity of EDA (Eq.~\eqref{eq:virtual_trajectory_summations}), 
the end-effector can reach $\mathbf{p}_{3}$ by superimposing another submovement $\mathbf{p}_{0,sub2}(t)$, without modifying $\mathbf{p}_{0,sub1}(t)$:
\begin{align}\label{eq:sequencing_two_submovements}
    \mathbf{p}_0(t) &= \mathbf{p}_{0,sub1}(t) + \mathbf{p}_{0,sub2}(t) \\
   \mathbf{p}_{0,sub2}(t) &= 
    \begin{cases}
      \mathbf{0}  &  \quad 0 \le t < t_g \\
      (\mathbf{p}_{3} - \mathbf{p}_{2}) f_{\sigma,2}(t) & \quad t_g \le t < t_g + T_2 \\
      \mathbf{p}_{3} & \quad t_g + T_2 \le t
    \end{cases} \nonumber
\end{align}
In these equations, $f_{\sigma,2}(t)$ is a second submovement starting at time $t_g$ with duration $T_2$.
With this new $\mathbf{p}_0(t)$ combined with the two submovements $\mathbf{p}_{0,sub1}(t)$ and $\mathbf{p}_{0,sub2}(t)$, the convergence of $\mathbf{p}(t)$ to the new goal $\mathbf{p}_3$ is achieved.

\subsection{Generating a Combination of Discrete and Rhythmic Movements }\label{subsec:sub_with_oscillation}
Consider a goal-directed discrete movement from initial position $\mathbf{p}_{1}\in\mathbb{R}^3$ to goal location $\mathbf{p}_{2}\in\mathbb{R}^3$. 
As discussed in Section \ref{subsec:sequence_of_discrete_movements}, this movement can be achieved by a single submovement $\mathbf{p}_{0,sub}(t)$ with amplitude $\mathbf{p}_2-\mathbf{p}_1$.
Our goal is to overlap a rhythmic movement onto this goal-directed discrete movement. 
This can be achieved by direct summation of an oscillation $\mathbf{p}_{0, osc}(t)$ (Eq.~\eqref{eq:EDA_oscillation}) onto $\mathbf{p}_{0,sub}(t)$:
\begin{equation}\label{eq:direct_sum_sub_and_osc}
    \mathbf{p}_0(t) = \mathbf{p}_{0,sub}(t) + \mathbf{p}_{0, osc}(t) 
\end{equation}

\begin{figure*}
    \centering
    \includegraphics[trim={0.0cm 3.5cm 0.0cm 0.0cm}, width=0.88\textwidth, clip, page=1]{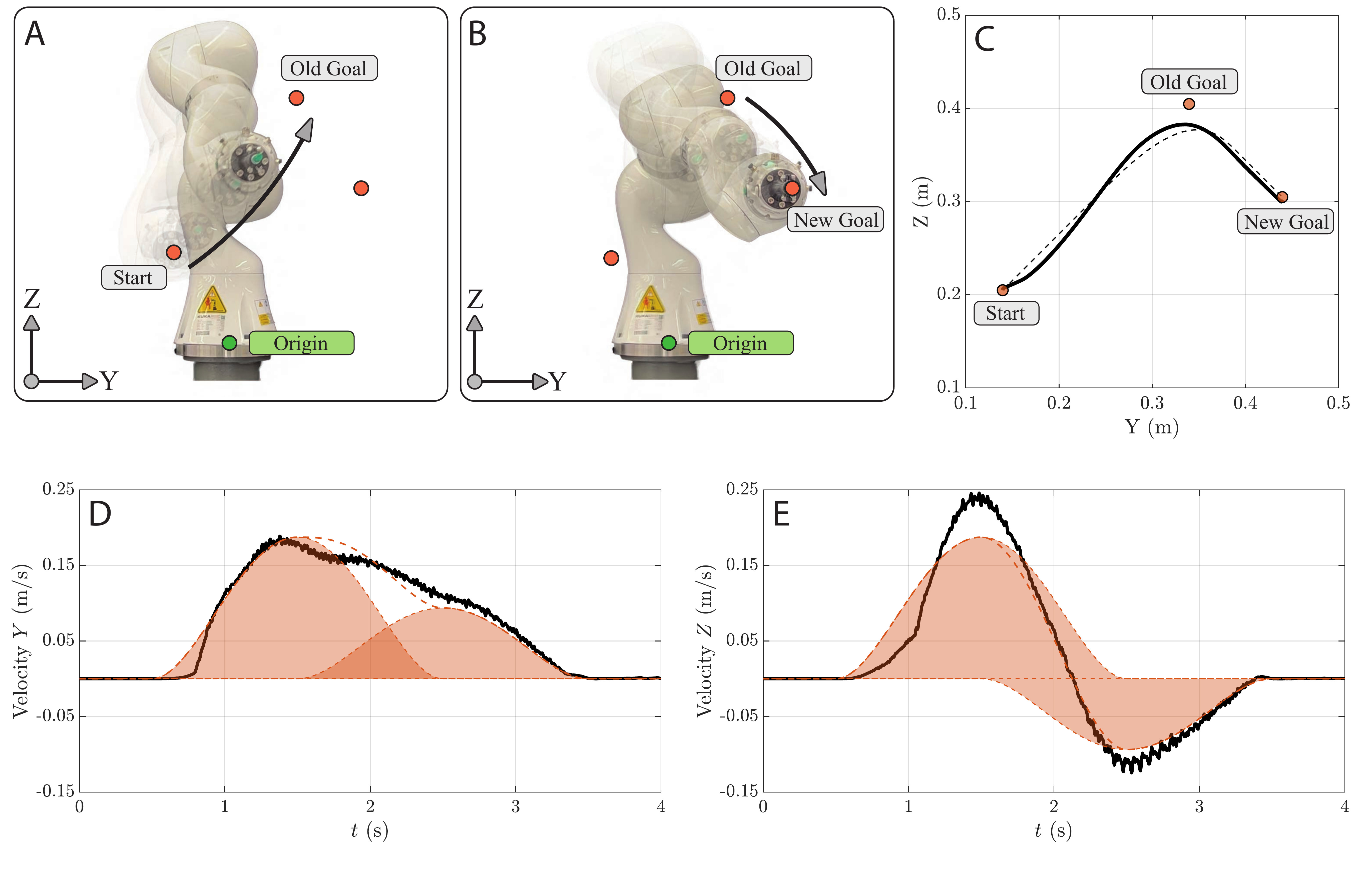}
    \caption{A sequence of discrete movements using a KUKA LBR iiwa14. (A, B) Time-frames of the robot movement towards the (A) original (old) and (B) new goal location. Start $\mathbf{p}_1$, original goal $\mathbf{p}_2$ and new goal $\mathbf{p}_3$ are depicted as orange markers. The origin of the robot's coordinate frame is attached at the robot base, depicted as a green marker. (C) The end-effector trajectory $\mathbf{p}(t)$ (black filled line) and the virtual trajectory (black dashed  line) $\mathbf{p}_0(t)$ depicted on the $YZ$-plane. (D, E) Time $t$ vs. end-effector velocity $\dot{\mathbf{p}}(t)$ along the (D) $Y$-coordinate and (E) $Z$-coordinate. Black filled lines show the end-effector velocity, which was derived by a first-order finite difference of $\mathbf{p}(t)$ with a sampling interval of 3ms. The two unimodal speed profiles filled in orange depict the two submovements $\mathbf{p}_{0,sub1}(t)$ (left) and $\mathbf{p}_{0,sub2}(t)$ (right) (Eq.~\eqref{eq:sequencing_two_submovements}). As shown in (D) and (E), the second submovement is directly superimposed, without any modification of the first submovement. Parameters of the submovements (Section \ref{subsec:experiment_sequence_movements}): $\mathbf{p}_1=[0.6735, 0.1396, 0.2048]$m, $\mathbf{p}_2=[0.6735, 0.3396, 0.4048]$m, $\mathbf{p}_3=[0.6735, 0.4396, 0.3048]$m, $T_1=T_2=2.0$s, $t_i=0.5$s, $t_g=1.5$s. } \label{fig:sequence_of_discrete_movements}  
\end{figure*}

\subsection{Drawing and Erasing Task}\label{subsec:drawing_and_erasing}
Consider a task of teaching a robot to draw a demonstrated trajectory $\mathbf{p}^{(d)}(t)\in\mathbb{R}^{2}$ on a table.
Without loss of generality, assume that the drawing table resides on a horizontal $XY$-plane.
After drawing $\mathbf{p}^{(d)}(t)$, the robot retraces $\mathbf{p}^{(d)}(t)$ backwards, with an additional oscillatory movement to erase the trajectory. 

For this, one can combine Imitation Learning with the kinematic modularity of EDA, where an oscillatory movement is directly superimposed (Section \ref{subsec:sub_with_oscillation}) onto the trajectory learned by DMP.
Imitation Learning with a two-dimensional DMP can be used to generate $\mathbf{p}_{0,dmp}(t)\equiv \mathbf{p}^{(d)}(t)$ and this trajectory was used as the $X$-, $Y$-coordinates of the virtual trajectory $\mathbf{p}_{0}(t)$ to draw $\mathbf{p}^{(d)}(t)$.

Once $\mathbf{p}^{(d)}(t)$ is drawn on the plane by $\mathbf{p}_{0,dmp}(t)$, the drawing can be erased by simply superimposing an oscillation $\mathbf{p}_{0, osc}(t)$ on a time-reversed trajectory of $\mathbf{p}_{0,dmp}(t)$:
\begin{equation*}
    \mathbf{p}_0(t) = 
    \begin{cases}
      \mathbf{p}_{0, dmp}(T-t) + \mathbf{p}_{0, osc}(t) &  \quad 0 \le t < T \\
      \mathbf{p}_{0, dmp}(0) + \mathbf{p}_{0, osc}(t) & \quad T \le t
    \end{cases}
\end{equation*}
In this equation, $T\in\mathbb{R}_{>0}$ is the duration of $\mathbf{p}^{(d)}(t)$.

Note that the $Z$-coordinates of $\mathbf{p}_0(t)$, $p_{0,z}(t)$, is not learned through Imitation Learning. Instead, an appropriate value for $p_{0,z}(t)$ must be chosen to remain in contact with the table.
To elaborate, consider the $Z$-coordinate of the drawing table to be $h_z$. 
The pen, extending from the robot's end-effector and facing downward in the $-Z$ direction, requires setting $p_{0,z}(t)$ to be lower than the drawing plane $h_z$ by an offset $\epsilon >0$, i.e., $p_{0,z}(t)=h_z-\epsilon$. This offset is determined based on the desired contact force. Furthermore, the orientation of the pen, crucial for drawing and erasing tasks, are maintained by setting a constant $\mathbf{R}_0$ of the controller (Eq.~\eqref{eq:impedances}). 
It is important to underline that the selected task not only is of interest to our modular claim (as it combines different kinematic actions as well as managing physical interaction), but it also represents a common scenario for multiple real-world scenarios such as polishing, cleaning, sanding, grinding, or even writing.

\section{Experimental Results}
For the robot experiment, a KUKA LBR iiwa14, with seven torque-actuated DOFs (i.e., $n=7$), was utilized. For control, KUKA's Fast Robot Interface (FRI) was used. The built-in gravity compensation was activated for all three tasks. For the impedance parameters $\mathbf{B}_q$, $\mathbf{K}_r$ and $\mathbf{B}_r$ (Eq.~\eqref{eq:impedances}), identical values were applied across all tasks. 
The impedance values were set as follows: $\mathbf{B}_q=1.0\mathbb{I}_7$ N$\cdot$m$\cdot$s/rad, where $\mathbb{I}_n\in\mathbb{R}^{n\times n}$ denotes an identity matrix; $\mathbf{K}_r=50\mathbb{I}_3$ N$\cdot$m/rad, $\mathbf{B}_r=5\mathbb{I}_3$ N$\cdot$m$\cdot$s/rad.
The robot's configuration $\mathbf{q}(t)$ was directly accessed through the FRI interface, and $\dot{\mathbf{q}}(t)$ was derived using a first-order finite difference of $\mathbf{q}(t)$ with a 3ms time step.
The Forward Kinematics Map for deriving $\mathbf{p}(t)$, $\mathbf{R}(t)$, and the Jacobian matrices $\mathbf{J}_p(\mathbf{q})$, $\mathbf{J}_r(\mathbf{q})$ was calculated with the Exp[licit]\textsuperscript{TM}-FRI Library.\footnote[2]{Github repository: https://github.com/explicit-robotics/Explicit-FRI}

\subsection{Generating a Sequence of Discrete Movements }\label{subsec:experiment_sequence_movements}
For the basis function of the two submovements $\mathbf{p}_{0,sub1}(t)$ and $\mathbf{p}_{0,sub2}(t)$, a minimum-jerk trajectory was used \cite{hogan1987moving}:
\begin{align*}
        f_{\sigma,1}(t) &= 10\Big( \frac{t-t_i}{T_1}\Big)^3 - 15\Big( \frac{t-t_i}{T_1}\Big)^4 + 6\Big( \frac{t-t_i}{T_1}\Big)^5 \\
        f_{\sigma,2}(t) &= 10\Big( \frac{t-t_g}{T_2}\Big)^3 - 15\Big( \frac{t-t_g}{T_2}\Big)^4 + 6\Big( \frac{t-t_g}{T_2}\Big)^5        
\end{align*}
The values $\mathbf{K}_p$, $\mathbf{B}_p$ (Eq.~\eqref{eq:impedances}) were $800\mathbb{I}_3$ N/m, $80\mathbb{I}_3$ N$\cdot$s/m, respectively. 
For the orientation, $\mathbf{R}_0$ was set to be constant.

\begin{figure*}
    \centering
    \includegraphics[trim={0.0cm 0.5cm 0.0cm 0.0cm}, width=0.88\textwidth, clip, page=1]{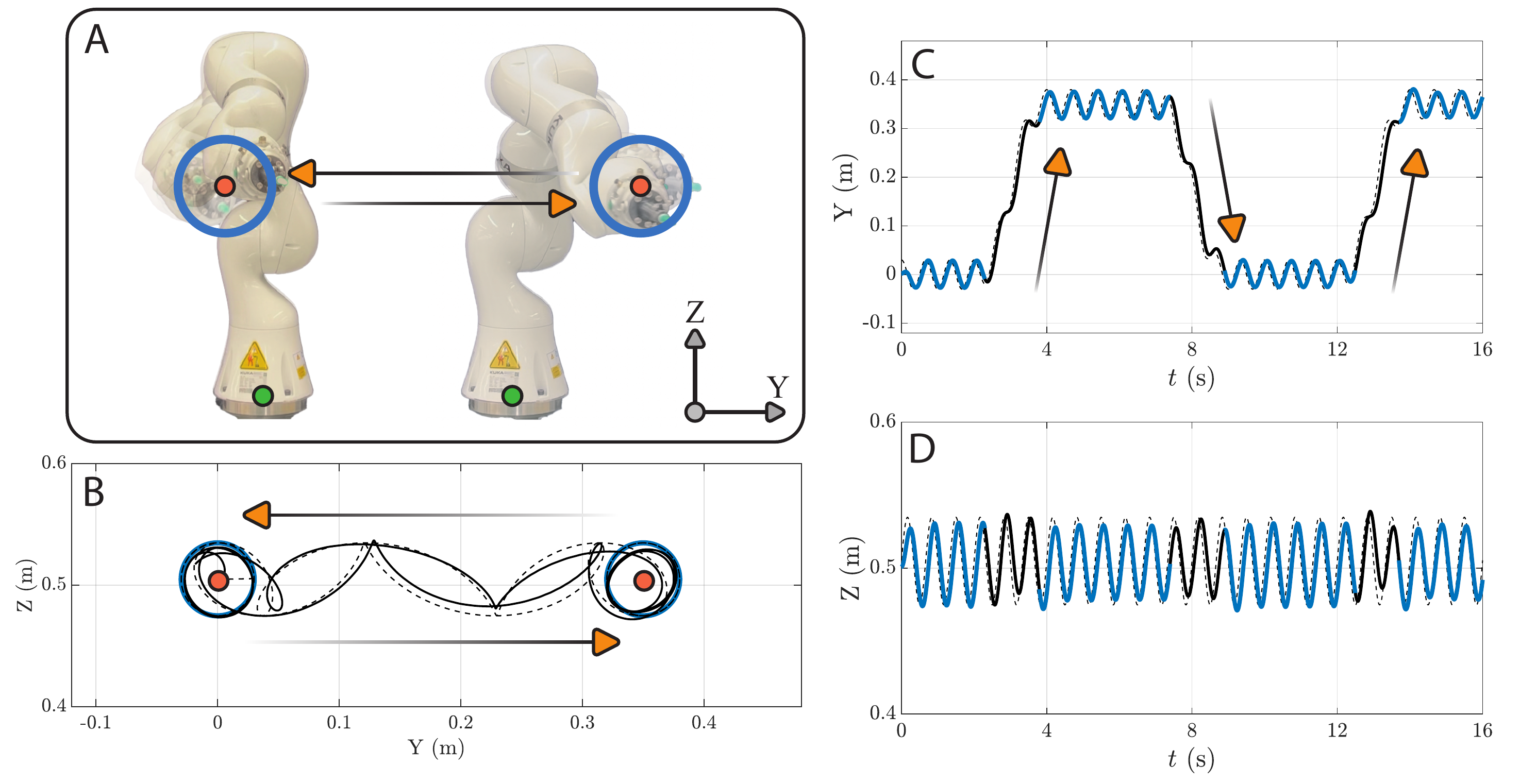}
    \caption{A combination of discrete and rhythmic movements using a KUKA LBR iiwa14. (A) Elements of the robot movement. The origin of the robot's coordinate frame is attached at the robot base, depicted as green marker. Orange markers depict $\mathbf{p}_1$ (left) and $\mathbf{p}_2$ (right) of $\mathbf{p}_{0,sub}(t)$. Blue line depicts $\mathbf{p}_{0,osc}(t)$ (Eq.~\eqref{eq:direct_sum_sub_and_osc}, \eqref{eq:experiment_oscillation}). (B) The end-effector trajectory $\mathbf{p}(t)$ (black filled line) and the virtual trajectory (black dashed line) $\mathbf{p}_0(t)$ depicted on the $YZ$-plane. Multiple submovements that move between $\mathbf{p}_1$ and $\mathbf{p}_2$ were generated. (C, D) Time $t$ vs. end-effector trajectory $\mathbf{p}(t)$ along (C) $Y$-coordinate and (D) $Z$-coordinate. Black dashed lines depict $\mathbf{p}_0(t)$. Blue lines highlight the duration of a movement without any discrete movement. Parameters of submovement and oscillation (Eq.~\eqref{eq:experiment_oscillation}): $\mathbf{p}_1=[0.5735, 0.0, 0.5048]$m, $\mathbf{p}_2=[0.5735, 0.35, 0.5048]$m, $T_1=1.5$s, $r=0.03$m, $\omega_0=3\pi$rad/s.} \label{fig:combination_of_discrete_and_rhythmic}      
\end{figure*}

The results are shown in Figure \ref{fig:sequence_of_discrete_movements}. 
With the proposed approach, a convergence of $\mathbf{p}(t)$ to the new goal location $\mathbf{p}_3$ was achieved (Figure \ref{fig:sequence_of_discrete_movements}A, \ref{fig:sequence_of_discrete_movements}B) by simply superimposing a second submovement onto the first one (Figure \ref{fig:sequence_of_discrete_movements}C, \ref{fig:sequence_of_discrete_movements}D, \ref{fig:sequence_of_discrete_movements}E). 
Note that the task was achieved without any modification of the first submovement.
Hence, the robot could adapt its movement to reach the new destination without the need for real-time modification of the initiated movements. 
This flexibility of the approach is advantageous in scenarios where quick adaptation is necessary to reach a new goal location.
Note that the simplicity of this approach is not guaranteed for other methods.
For instance, using DMP, the task requires online modification of the initiated movement \cite{nemec2012action}, which may introduce practical difficulties for implementation. 

\begin{figure*}
    \centering
    \includegraphics[trim={0.0cm 38.0cm 0.0cm 0.0cm}, width=0.93\textwidth, clip, page=1]{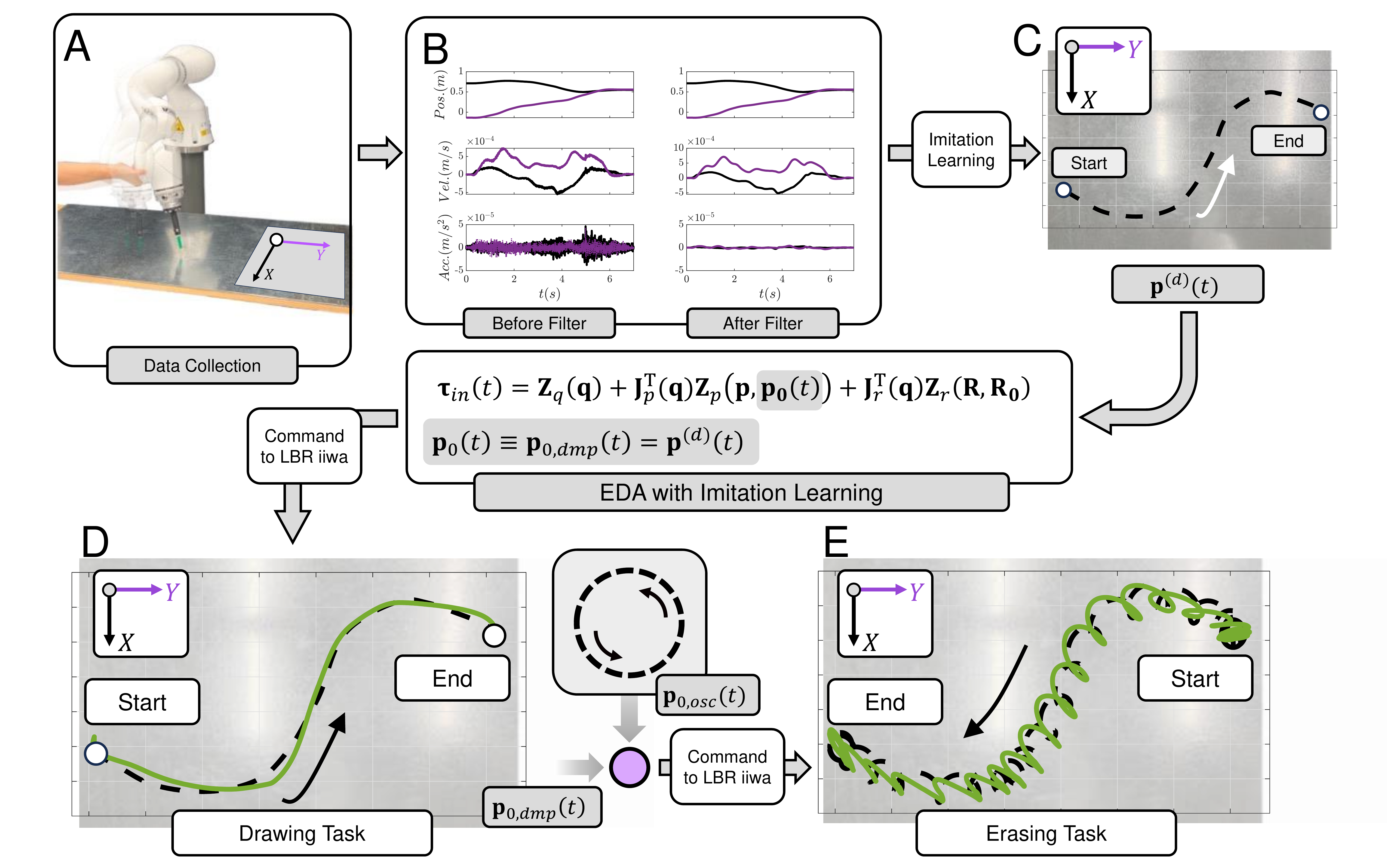}
    \caption{The drawing and erasing task using a KUKA LBR iiwa. A green pen was used for the drawing. (A) Data collection of human-demonstrated $\mathbf{p}^{(d)}(t)$ which was to be drawn (Section \ref{subsec:drawing_and_erasing}). The end-effector trajectory along $X$- and $Y$-coordinates were collected. (B) Time $t$ vs. $X$-coordinate (black line) and $Y$-coordinate (purple line) of $\mathbf{p}^{(d)}(t)$ (top row), $\dot{\mathbf{p}}^{(d)}(t)$ (middle row), $\ddot{\mathbf{p}}^{(d)}(t)$ (bottom row). With a sampling rate of 333Hz, a first-order finite difference method was used to calculate the velocity and acceleration (left column). These trajectories were Gaussian filtered (right column) using MATLAB's \texttt{smoothdata} function with a time window size of 165ms. (C) The resulting trajectory $\mathbf{p}^{(d)}(t)$ (black dashed line) generated with Imitation learning. (D) The drawing task was achieved by setting $\mathbf{p}_0(t)$ as $\mathbf{p}^{(d)}(t)$. The black dashed line depicts $\mathbf{p}_{0,dmp}(t)$ (or $\mathbf{p}^{(d)}(t)$), the green line depicts $\mathbf{p}(t)$. (E) The erasing task was achieved by superimposing an oscillation $\mathbf{p}_{0,osc}(t)$ onto a time-reversed $\mathbf{p}_{0,dmp}(t)$. The green pen was replaced by a rectangular eraser. The green line depicts $\mathbf{p}(t)$. For (C, D, E), trajectories were plotted in MATLAB and overlapped onto the drawing/erasing table. Parameters of DMP: $\alpha_z=1000$, $\beta_z=250$, $N=100$, $P=2331$, $\tau=7$, $c_i = \exp( -\alpha_s (i - 1)/ ( N-1) )$, $h_i=1/(c_{i+1} - c_i)^2$ for $i\in[1,2,\cdots, N-1]$, $c_{N}=\exp{(-\alpha_s)}$, $h_{N}=h_{N-1}$. Parameters of oscillation: $r=0.03$m, $\omega_0=2\pi$rad/s.} \label{fig:drawing_and_erasing}.     
\end{figure*}

\subsection{Generating a Combination of Discrete and Rhythmic Movements }\label{subsec:gen_comb_discrete_and_rhythmic}
For the submovement, $\mathbf{p}_{0,sub}(t)$ with minimum-jerk trajectory was employed.
For the oscillation, a circular trajectory residing on the $YZ$-plane was used:
\begin{equation}\label{eq:experiment_oscillation}
    \mathbf{p}_{0,osc}(t) = r [ 0, \cos(\omega_0 t), \sin(\omega_0 t ) ]
\end{equation}
In this equation, $r$ and $\omega_0$ are the radius and angular velocity of the circular trajectory, respectively.
The values of the impedance parameters were identical to those in Section \ref{subsec:experiment_sequence_movements}, i.e., $\mathbf{K}_p=800\mathbb{I}_3$ N/m and $\mathbf{B}_{p}=80\mathbb{I}_3$ N$\cdot$s/m, respectively (Eq.~\eqref{eq:impedances}).
For the orientation, $\mathbf{R}_0$ was set to be constant.

The results are shown in Figure \ref{fig:combination_of_discrete_and_rhythmic}.
The proposed approach enabled a combination of discrete and rhythmic movements of the robot's end-effector (Figure \ref{fig:combination_of_discrete_and_rhythmic}A) through the superposition of submovement and oscillation (Figure \ref{fig:combination_of_discrete_and_rhythmic}B, \ref{fig:combination_of_discrete_and_rhythmic}C, \ref{fig:combination_of_discrete_and_rhythmic}D).
It is important to note that achieving a direct combination of both discrete and rhythmic movements presents a challenge for DMP methodologies, which typically treat these movement types separately \cite{degallier2006movement,degallier2007hand} (Section \ref{subsec:DMP_and_IL}).

\subsection{Drawing and Erasing Task}
For Imitation Learning of $\mathbf{p}^{(d)}(t)$, the human-demonstrated data points of $\mathbf{p}^{(d)}(t)$ were collected with a sampling rate of 333Hz. 
The velocity $\dot{\mathbf{p}}^{(d)}(t)$ and acceleration $\ddot{\mathbf{p}}^{(d)}(t)$ for Imitation Learning were derived by first-order finite difference of $\mathbf{p}^{(d)}(t)$ with Gaussian filtering. 
For the filtering, MATLAB's $\texttt{smoothdata}$ function with a time window size 165ms was used. 
With these filtered data, Linear Least Square Regression was used (Section \ref{subsubsec:imitation_learning}).
For the drawing and erasing tasks, the impedance parameters $\mathbf{K}_p=400\mathbb{I}_3$ N/m (respectively $800\mathbb{I}_3$ N/m) and $\mathbf{B}_p=40\mathbb{I}_3$ N$\cdot$s/m (respectively $80\mathbb{I}_3$ N$\cdot$s/m) were used.
Additionally, for the erasing task, an oscillation $\mathbf{p}_{0,osc}(t)$, as described in Eq.~\eqref{eq:experiment_oscillation} was implemented on the $XY$-plane, instead of the $YZ$-plane. 

The results are shown in Figure \ref{fig:drawing_and_erasing}, illustrating the entire process of generating $\mathbf{p}_0(t)$ for the drawing and erasing tasks. By merging Imitation Learning with EDA (Figure \ref{fig:drawing_and_erasing}A, \ref{fig:drawing_and_erasing}B, \ref{fig:drawing_and_erasing}C), the drawing (Figure \ref{fig:drawing_and_erasing}D) and erasing tasks (Figure \ref{fig:drawing_and_erasing}E) were successfully achieved. 
It is worth emphasizing that the key to this approach is the combination of Imitation Learning and the modular property of EDA.
The trajectory $\mathbf{q}^{(d)}(t)\equiv \mathbf{q}_{0,dmp}$ was separately learned with Imitation Learning and directly combined with an oscillation $\mathbf{p}_{0,osc}(t)$. 
With modest parameter tuning (i.e., changing the angular velocity $\omega_0$), trajectory $\mathbf{p}_{0,osc}(t)$ used in task \ref{subsec:gen_comb_discrete_and_rhythmic} was simply reused. 
Using appropriate values of mechanical impedances, stability against contact and physical interaction was achieved for both drawing and erasing tasks.
Note that the simplicity of the approach is not immediately apparent when solely using DMP.
To generate discrete and rhythmic movements with DMP requires different types of canonical system and nonlinear forcing term \cite{ijspeert2013dynamical,saveriano2023dynamic}.
Hence, one cannot directly combine these two movements. 
Moreover, even if an additional method to merge these movements were devised, mapping this task-space trajectory to joint-space commands would require additional consideration (e.g., managing kinematic redundancy).
By merging EDA with DMP, these problems are avoided. 

\section{Discussion, Limitations and Conclusion}\label{sec:discussion_and_conclusion}
Thanks to the kinematic modularity of EDA, the two tasks of sequencing discrete movements and combining discrete and rhythmic movements were greatly simplified.
For the former, the subsequent movement was directly superimposed onto the previous movement without modifying the first submovement.
For the latter, the discrete and rhythmic movements were planned separately and then directly superimposed. 
The authors want to emphasize that the simplicity of this approach is significant and non-trivial. 
For instance, with DMP, the sequencing task requires  additional dynamics to reach the goal $\mathbf{x}_g$. 
This can lead to practical challenges in real-world robot implementation.
In contrast to EDA, for combination tasks, DMP treats discrete and rhythmic movements separately \cite{degallier2006movement,degallier2007hand,degallier2008modular,ijspeert2013dynamical}, which thereby prevents  direct superposition of the two types of movements generated (or learned) by DMP.

The robot demonstration of the drawing and erasing task illustrated how to get the best of both motor-primitive approaches.
DMP and Imitation Learning provided a rigorous mathematical framework to generate (or learn) trajectories of arbitrary complexity. 
EDA and its Norton equivalent network structure provided a modular framework for robot control.
Using Imitation Learning to learn the virtual trajectory of EDA, a modular learning strategy for robot control was implemented. 

Merging the methods of EDA with DMP preserved the kinematic modularity of EDA and combined it with favorable behavior during physical interaction. 
This facilitated the drawing and erasing tasks, which involved physical contact. 
The key to this approach is the compliant robot behavior that emerges from mechanical impedance, a feature that might be challenging to replicate using position-actuated robots \cite{hogan2022contact}.
However, the compliance provided by mechanical impedance led to a non-negligible tracking error between the virtual and actual end-effector trajectories. 
This issue could be mitigated to some extent by increasing the impedance values in the direction of the desired motion. 

The modular approach with torque-actuated robots, as presented in this paper, offers significant advantages over position-actuated robots. Utilizing EDA with torque-actuated robots eliminates issues related to inverse kinematics, kinematic singularity, and redundancy. Conversely, employing position-actuated robots for task-space control requires additional methods, such as damped least-square methods \cite{buss2005selectively} and generalized pseudo-inverse matrices \cite{nakamura1986inverse}, which in turn complicates the controller design. However, if only position-actuated robots are available, the torque command can be mapped to accelerations using the Forward Dynamics model (i.e., mapping from torque to position command) of the robot.

As discussed, a key objective of EDA is to find appropriate choices of virtual trajectory and the corresponding mechanical impedance. 
The former property was addressed in this paper through the use of DMP, resulting in modular Imitation Learning. 
However, selecting appropriate values for mechanical impedance may not be straightforword; the presented values were identified by trial-and-error. 
A systematic method to choose (or learn) the impedance parameters is an avenue of future research \cite{buchli2011learning}. 

In conclusion, by integrating EDA with DMP's Imitation Learning, we have established a modular learning strategy for robot control. This approach significantly simplifies the generation of a broad spectrum of robot actions. 




\bibliographystyle{IEEEtran}
\bibliography{references.bib}

\addtolength{\textheight}{-12cm}   

\end{document}